\documentclass[11pt]{TinLatex} 

\usepackage{graphicx}
\usepackage{epstopdf}

\usepackage{amsmath,amsxtra,amssymb,latexsym, amscd}

\usepackage[mathscr]{eucal}

\usepackage{hyperref}
\hypersetup{
 colorlinks = true,
 linkcolor  = red,
 citecolor  = red,
 urlcolor   = black
}

\usepackage{xspace}

\def\centerbmp#1#2#3{\vskip#2\relax\centerline{\hbox to#1{\special
  {bmp:#3 x=#1, y=#2}\hfil}}}

\def\centereps#1#2#3{\vskip#2\relax\centerline{\hbox to#1{\special
  {eps:#3 x=#1, y=#2}\hfil}}}

\def\centerwmf#1#2#3{\vskip#2\relax\centerline{\hbox to#1{\special
  {wmf:#3 x=#1, y=#2}\hfil}}}

\def\centerps#1#2#3{\vskip#2\relax\centerline{\hbox to#1{\special
  {ps:#3 x=#1, y=#2}\hfil}}}

\input dih.tex

\font\it=cmti12 at 11pt
\font\bf=cmbx12 at 11pt

\advance\hoffset0.4cm
\font\bfh=cmssbx10 scaled 1400
\font\bfl=cmssbx10 scaled \magstep1

\font\cn=cmr10
\def\ed{\end{document}}
\def\vuong{\raise-0.16cm\hbox{$^\blacksquare$}}

\makeatletter
\DeclareRobustCommand\onedot{\futurelet\@let@token\@onedot}
\def\@onedot{\ifx\@let@token.\else.\null\fi\xspace}

\makeatother

\usepackage{multirow}
\usepackage{enumitem}
\usepackage[T1]{fontenc}
\usepackage[utf8]{inputenc}
\usepackage{lmodern}

\usepackage[utf8]{vietnam}
\usepackage{subfigure}
\usepackage{tabularx}
\usepackage{booktabs}
\usepackage{amssymb}
\usepackage{booktabs}
\usepackage[ruled,linesnumbered]{algorithm2e}
	\SetAlgorithmName{Algorithm}{algorithm}{}
\usepackage{amsmath}

\NewDocumentCommand{\MYref}{O{black}mo}{%
  \begingroup
  \hypersetup{linkcolor=#1}%
  \ref{#2}%
  \IfValueT{#3}{%
    \color{#1}{#3}%
  }%
  \endgroup
}% %NEW COMMAND ALTERNATIVE TO REF

\usepackage{caption} 
\usepackage{cleveref}
\usepackage{color,soul}
\usepackage{pagecolor}
\usepackage{amsthm}
\usepackage{amsfonts}
\usepackage{amssymb}
\usepackage{amsmath}

\usepackage{orcidlink}
\usepackage{algpseudocode}
\usepackage{indentfirst}
\usepackage[english]{babel}
\usepackage[table]{xcolor}

\usepackage{graphicx}

\begin{document}
\makeatletter	   % `@' is now a normal "letter' for LaTeX
\setcounter{page}{1}
% \refnameenewcommand{\ps@plain}{%
% 	\renewcommand{\@oddhead}{\hfill\begin{tabular}{r}
% 			\small{\emph{Journal of Computer Science and Cybernetics, V.40, N.3 (2025), 1-}} \\
% 			\footnotesize{DOI: 10.15625/1813-9663/19558

% 		\end{tabular}
% 	}
	
% 	\renewcommand{\@evenhead}{\@oddhead}
% 	\renewcommand{\@oddfoot}{
% 		%\small\hfill{\hskip7.1cm \copyright\ 2024 Vietnam  Academy of Science \& Technology}
% 		\renewcommand{\@evenfoot}{\@oddfoot}
% 	}
% } }

%Lenh duoi thay doi do dai cua duong ngang
%\usepackage[flushmargin]{footmisc} - PHẢI MỞ LỆNH NÀY
\def\footnoterule{\kern-3\p@
	\hrule \@width 1.32in \kern 2.6\p@} % the \hrule is .4pt high
%Thay doi indent cua footnote
%\renewcommand{\footnotesep}{2cm}
% \renewcommand{\footnotemargin}{0em}
\newcommand{\fn}[1]{\footnotetext{\hspace{-6mm}#1}}
\setlength{\skip\footins}{8mm}

\makeatother   % `@' is restored as a "non-letter" character 
    
%%%%%%%%%%%%%%%%%%%%%%%%
%Title & Authors
%%%%%%%%%%%%%%%%%%%%%%%%

%=========================================================================
\title{Digital FAST: An AI-Driven Multimodal Framework for Rapid and Early Stroke Screening}
% \subtitle{SUBTITLE}
\author{
	{\cn NGOC-KHAI HOANG$^1$, THI-NHU-MAI-NGUYEN$^3$, HUY-HIEU PHAM$^{1,2,*}$}
	\vskip.5cm
	{
    \it $^1$VinUni-Illinois Smart Health Center, VinUniversity, Hanoi 100000, Vietnam\\
        \it $^2$College of Engineering and Computer Science, VinUniversity, Hanoi 100000, Vietnam\\
        \it $^3$ Study Program Medical Informatics, \textit{{\fontencoding{T1}\selectfont Universität zu Lübeck}}, Germany \\
        \it *Email: hieu.ph@vinuni.edu.vn \\     
% \fn{*Corresponding author. Huy-Hieu Pham (email: hieu.ph@vinuni.edu.vn) }
% \fn{\hspace{1.7mm}{\it E-mail addresses}:
% \href{hieu.ph@vinuni.edu.vn}{hieu.ph@vinuni.edu.vn} (Huy-Hieu Pham); 
% \href{mailto:anonymous@ioit.ac.vn}{anonymous@tnu.edu.vn} (AUTHOR NAME(S) 2).
% }
}
}
\maketitle
\renewcommand\refname{\normalsize \centerline{ REFERENCES}}
% Set to use the "plain" pagestyle
\pagestyle{plain}
\pagestyle{myheadings}
% \markboth{\footnotesize \chu \uppercase{NGOC-KHAI HOANG} {\it et al.}}
% \markboth
% {\footnotesize \uppercase{NGOC-KHAI HOANG et al.}}
% {\footnotesize \uppercase{NGOC-KHAI HOANG et al.}}
% {\footnotesize  \chu \uppercase{Digital FAST: An AI-Driven Multimodal Framework for Rapid and Early Stroke Screening }}% Maximun 8 words 
% {	\vspace{-.4cm}
% 	%\hspace{13.25cm}{\includegraphics[scale=.2]{crosscheck.pdf}}%\vspace{-.5cm}
% 	\hspace{13.04cm}{\includegraphics[scale=.05]{cross_check.PNG}
% 	}
% }
%=========================================================================

%=========================================================================
\begin{abstract} {
Early identification of stroke symptoms is essential for enabling timely intervention and improving patient outcomes, particularly in prehospital settings. This study presents a fast, non-invasive multimodal deep learning framework for automatic binary stroke screening based on data collected during the F.A.S.T. assessment. The proposed approach integrates complementary information from facial expressions, speech signals, and upper-body movements to enhance diagnostic robustness. Facial dynamics are represented using landmark based features and modeled with a Transformer architecture to capture temporal dependencies. Speech signals are converted into mel spectrograms and processed using an Audio Spectrogram Transformer, while upper-body pose sequences are analyzed with an MLP-Mixer network to model spatiotemporal motion patterns. The extracted modality specific representations are combined through an attention-based fusion mechanism to effectively learn cross modal interactions. Experiments conducted on a self-collected dataset of 222 videos from 37 subjects demonstrate that the proposed multimodal model consistently outperforms unimodal baselines, achieving 95.83\% accuracy and a 96.00\% F1-score. The model attains a strong balance between sensitivity and specificity and successfully detects all stroke cases in the test set. These results highlight the potential of multimodal learning and transfer learning for early stroke screening, while emphasizing the need for larger, clinically representative datasets to support reliable real-world deployment. }
%=========================================================================

\end{abstract}

%%%%%%%%%%%%%%%%%%%%%%%%
%Main text
%%%%%%%%%%%%%%%%%%%%%%%%

%=========================================================================
\section{INTRODUCTION}
Stroke refers to a group of cerebrovascular disorders caused by the formation of blood clots or hemorrhagic events that obstruct or rupture blood vessels supplying the brain. Among these conditions, acute ischemic stroke is one of the leading causes of long-term disability and mortality worldwide\cite{johnston2009global, world2014guidelines}, with epidemiological studies estimating that approximately one in four individuals will experience a stroke during their lifetime \cite{feigin2025world, feigin2021global, johnson2019global}. Consequently, timely and appropriate intervention is of critical importance\cite{saver2006time}. Early detection of stroke is essential to enable prompt preventive measures and effective treatment.
Notably, when individuals exhibit early signs of stroke, bystanders often lack sufficient medical knowledge to recognize these symptoms as accurately as healthcare professionals\cite{harbison2003diagnostic}. This gap highlights the urgent need for automated methods capable of detecting stroke risk in real-world settings, particularly for non-expert users.
In the context of rapid digitalization, artificial intelligence (AI) has significantly impacted the healthcare domain. Deep learning models constitute the foundation of various applications\cite{litjens2019state}, including automated healthcare systems, clinical decision support, and medical diagnosis, where AI has, in certain cases, achieved performance superior to that of human specialists\cite{alowais2023revolutionizing, topol2019high}.
Multimodal learning approaches, which integrate information from multiple data sources, have demonstrated superior performance compared to models relying on a single modality\cite{benani2025multimodal, baltruvsaitis2018multimodal, ramachandram2017deep}. 
However, in light of the rapid digitalization of healthcare and the growing adoption of multimodal deep learning techniques, there remains a notable gap between methodological advances and their application to practical, high-impact clinical scenarios such as prehospital stroke screening. Although prior studies have demonstrated the advantages of integrating heterogeneous data sources\cite{acosta2022multimodal}, multimodal stroke detection particularly under subject independent and prehospital settings has not yet been sufficiently explored\cite{shurrab2024multimodal, heo2019machine}. Addressing this gap requires not only effective multimodal modeling strategies but also datasets and experimental settings that reflect real-world constraints.

Building upon this motivation, this study aims to advance multimodal stroke screening by jointly exploiting complementary cues from visual, acoustic, and motion based signals collected during the F.A.S.T. assessment\cite{does2007designing}. In particular, we focus on designing a practical and non-invasive solution suitable for prehospital scenarios, where rapid assessment, robustness to limited data, and reliable generalization across subjects are critical.
Our main contributions of this study are outlined as follows:
\begin{itemize}
    \item We construct a self-collected multimodal dataset for prehospital stroke screening, consisting of temporally synchronized facial videos, speech recordings, and upper-body pose data acquired during the F.A.S.T. assessment.
    \item We develop a novel multimodal deep learning framework that integrates Transformer-based encoders for facial and speech modalities with an MLP-Mixer architecture for pose representation, combined through an attention-based fusion mechanism to effectively capture cross-modal dependencies.
    \item We conduct extensive experimental evaluations and ablation studies under subject independent settings, demonstrating that the proposed multimodal approach consistently outperforms unimodal and pairwise fusion baselines while maintaining a robust balance between sensitivity and specificity.

\end{itemize}
The remainder of this paper is organized as follows. 
Section~\ref{Related Work} reviews related work on stroke detection and multimodal deep learning approaches. 
Section~\ref{Proposed Method} presents the proposed multimodal framework, including the modality specific encoders and the fusion strategy. 
Section~\ref{Dataset} describes the dataset and data acquisition process in detail. 
Section~\ref{Experiment} reports the experimental setup and quantitative performance results. 
Section~\ref{Ablation Study} provides ablation studies to analyze the contribution of individual modalities and fusion strategies. 
Section~\ref{Dicussion} discusses the experimental findings, limitations, and practical implications of the proposed method. 
Finally, Section~\ref{Conclusion} concludes the paper and outlines directions for future work.

\section{Related Work} \label{Related Work}
Recent advances in artificial intelligence have led to increasing interest in applying machine learning and deep learning techniques to stroke diagnosis and prognosis. A comprehensive systematic review by Shurrab et al.\cite{shurrab2024multimodal} systematically analyzed multimodal machine learning approaches for stroke related tasks, highlighting that most existing studies focus on neuroimaging modalities such as computed tomography (CT) and magnetic resonance imaging (MRI) for stroke detection, brain tissue segmentation, and outcome prediction. In addition, electronic health records and clinical data are frequently utilized to estimate stroke risk and predict functional outcomes. Despite the growing popularity of multimodal learning, the review also emphasized the limited availability of diverse multimodal datasets and the need for methods tailored to practical clinical and prehospital scenarios.

Beyond imaging-based approaches, several studies have explored non imaging indicators relevant to stroke detection, including facial abnormalities, speech impairments, and motor dysfunction. Wang et al.\cite{wang2024prediagnosis} proposed an ensemble convolutional neural network integrating Xception, ResNet50, VGG19, and EfficientNet-B1 to detect acute ischemic stroke from two-dimensional facial images. Their model achieved an AUC of 0.91 on a large dataset of stroke patients and matched controls, demonstrating the feasibility of facial image based stroke screening prior to CT examination.

To address ethical and privacy concerns associated with using raw facial videos, Cai et al.\cite{cai2025safetriage} introduced SafeTriage, a privacy preserving stroke triage framework that de-identifies facial videos while retaining diagnostically relevant motion patterns. Their approach employs a pretrained video motion transfer model to map facial motion onto synthetic identities, combined with a conditional generative model to mitigate distribution shifts. Experimental evaluations showed that the generated synthetic videos preserved stroke relevant facial dynamics while maintaining diagnostic accuracy and strong privacy protection.

Speech impairments, particularly dysarthria, have also been widely studied as indicators of neurological disorders including stroke. Mahum et al.\cite{mahum2025novel} proposed a Swin Transformer based framework for dysarthria recognition using mel spectrogram representations, enhancing the model’s ability to capture local articulatory patterns through specialized architectural modules. Their method achieved high recognition accuracy, demonstrating the effectiveness of transformer-based models for speech related neurological assessment. Similarly, Yang et al.\cite{yang2024feature} introduced a contrastive learning based feature extraction method for dysarthria detection, where pretrained models were used to learn discriminative representations between healthy and impaired speech before classification.

More recently, multimodal approaches combining visual, speech, and movement information have been investigated for stroke triage in acute and prehospital settings. Cai et al.\cite{cai2024m} proposed M3 Stroke, a mobile multimodal AI system integrating audio-visual data for emergency triage of mild to moderate stroke patients. Trained and evaluated on real patient data, their system significantly improved sensitivity and specificity compared to traditional triage methods, demonstrating the potential of multimodal AI for real-world deployment. Similarly, Ou et al.\cite{ou2025early} developed a multimodal deep learning framework based on the F.A.S.T. assessment, incorporating facial expressions, speech recordings, and limb movement data collected from emergency room patients. Their results showed that multimodal models outperformed unimodal variants, highlighting the benefit of jointly modeling heterogeneous patient signals.

% Although these studies demonstrate the promise of multimodal and non-invasive stroke screening, existing methods either focus on a limited set of modalities, rely on raw identity sensitive data, or are designed primarily for clinical rather than prehospital contexts. Motivated by these limitations, the present study aims to develop a privacy aware, multimodal deep learning system that integrates facial landmarks, speech signals, and upper body movement information for early stroke screening in prehospital environments.

% Although these studies demonstrate the promise of multimodal and non-invasive stroke screening, existing methods either focus on a limited set of modalities, rely on raw identity-sensitive data, or are designed primarily for clinical rather than prehospital contexts.

Further multimodal learning approaches have also been explored for rapid stroke assessment in emergency clinical settings. Yu et al.\cite{yu2020toward} proposed a multimodal deep learning framework designed to support rapid stroke diagnosis by emulating clinical screening protocols such as the Cincinnati Prehospital Stroke Scale (CPSS) and the Face Arm Speech Test (FAST). Their approach integrates facial motion analysis and speech assessment extracted from patient video recordings to automatically detect neurological deficits associated with stroke. Experiments
conducted on video data collected from real patients in emergency rooms demonstrated that the proposed system achieved diagnostic performance comparable to that of emergency physicians while enabling faster automated screening.

Building upon this direction, Cai et al.\cite{cai2022deepstroke} introduced DeepStroke, a multimodal adversarial deep learning framework for stroke screening in emergency room environments. The system jointly analyzes facial muscle coordination and speech impairments to detect subtle neurological abnormalities indicative of stroke.
By leveraging multimodal feature learning and adversarial training strategies, DeepStroke effectively captures complementary information from visual and acoustic modalities. Experimental results showed that the proposed framework achieved high sensitivity for stroke detection while maintaining practical screening time suitable for emergency clinical workflows.

Although these studies demonstrate the effectiveness of multimodal learning for stroke assessment, most existing systems are designed for controlled clinical environments and rely on raw facial video data, which may raise privacy concerns and limit their applicability in practical prehospital scenarios. Motivated by these limitations, the present study aims to develop a privacy-aware multimodal deep learning system that integrates facial landmarks, speech signals, and upper-body movement information for early stroke screening in prehospital environments.

\section{Proposed Method} \label{Proposed Method}

\begin{figure*}[ht]
    \centering
    \includegraphics[width=1\linewidth]{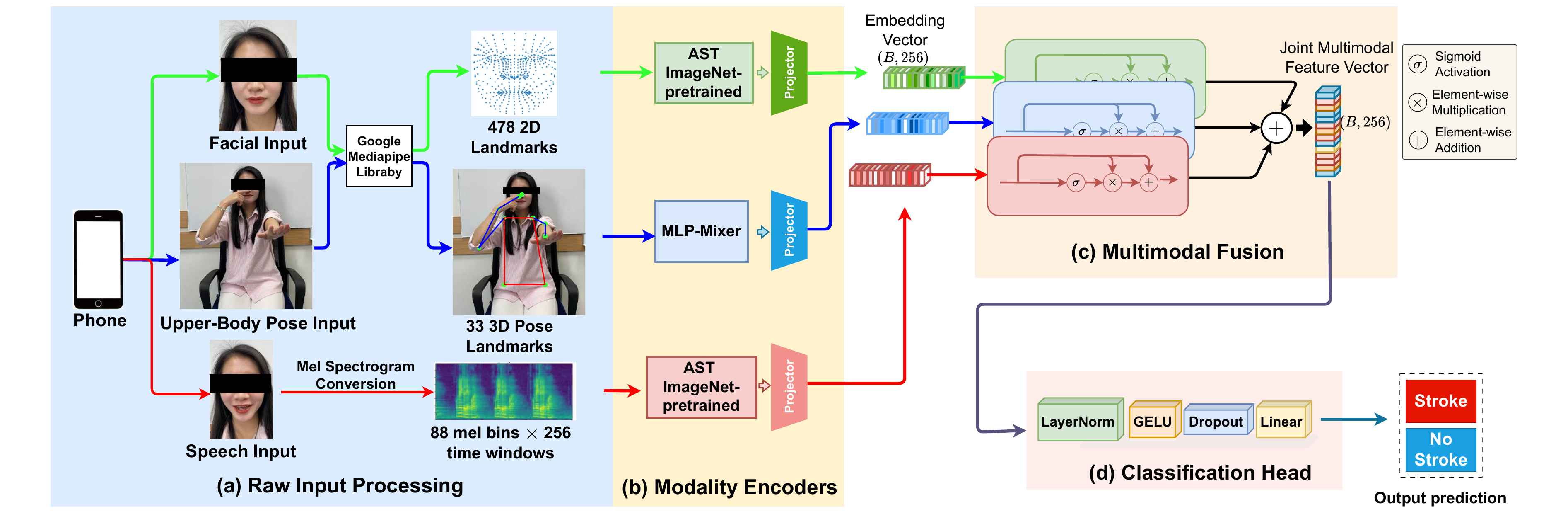}
    \caption{The proposed framework comprises four stages: (a) Raw Input Processing, which transforms raw visual and audio data into structured representations; (b) Modality Encoders, which extract high-level embeddings using modality-specific models; (c) Multimodal Fusion, which integrates the embeddings through a learnable fusion mechanism; and (d) Classification Head, which produces the final binary stroke prediction.}
    \label{Model}
\end{figure*}

We propose a multimodal deep learning framework that jointly leverages facial expressions, upper-body movements, and speech signals for automated stroke screening in prehospital settings, as illustrated in Fig.~\ref{Model}. The system takes as input facial videos, upper-body videos, and speech signals collected from participants during the F.A.S.T. assessment. These data streams are processed through modality-specific branches, each tailored to the characteristics of the corresponding modality, before being integrated at the feature representation level. 

\subsection{Raw Input Processing}

As illustrated in Fig.~\ref{Model}(a), the system processes three types of raw inputs: facial videos, upper-body videos, and speech signals collected during the F.A.S.T. assessment task.
In the visual branch, facial and upper-body videos are first processed using the Google MediaPipe library to extract landmark representations. For facial analysis, facial landmarks capture the geometric structure and asymmetry of critical regions such as the mouth and cheeks, which are directly associated with hallmark symptoms of stroke. In parallel, upper-body pose landmarks are employed to model the kinematics of the torso and limb movements, enabling the detection of motor abnormalities, such as impaired coordination during motor assessment tasks.

Landmarks are extracted on a frame-by-frame basis and organized into temporal sequences, yielding a sequential representation of motion dynamics over time. Specifically, we treat each frame as a token in a temporal sequence. For the face modality, each token is a flattened 2D landmark vector of dimension $D_f = 478 \times 2 = 956$; for the pose modality, each token is a flattened 3D vector of dimension $D_p = 33 \times 3 = 99$. A linear projection maps each per-frame token into the encoder embedding space, and positional embeddings are added to preserve temporal order. During batching, sequences are padded (or center-cropped if longer) along the temporal dimension to ensure consistent tensor shapes across samples.

For the auditory modality, speech signals are converted into mel-spectrogram representations using a short-time Fourier transform with sliding windows, followed by a transformation to the mel scale to emphasize perceptually relevant frequency bands. These spectrograms serve as the input to the speech encoder.

\subsection{Modality Encoders}

As shown in Fig.~\ref{Model}(b), each modality is processed using a dedicated encoder designed to capture modality-specific patterns.

In the first visual sub-branch, the facial landmark sequences are directly fed into an Audio Spectrogram Transformer (AST) model pretrained on ImageNet to extract spatio-temporal features. Applying AST to landmark-based representations allows the model to capture long-range temporal dependencies by formulating the motion sequence as a patch-based representation learning problem, analogous to image processing or time-frequency analysis.

In contrast, the upper-body pose landmark sequences are processed using an MLP-Mixer architecture to effectively model interactions along both spatial and temporal dimensions, which aligns well with the kinematic characteristics of body movements. Beyond its role as a lightweight feature extractor, the MLP-Mixer is particularly well suited for pose-based representations, as it explicitly disentangles token-mixing operations, corresponding to interactions among body joints, from channel-mixing operations, which capture temporal dynamics across frames.

This design enables efficient learning of structured geometric relationships and motion patterns without relying on complex attention mechanisms, making it robust to variable sequence lengths and noisy landmark estimates commonly encountered in prehospital data. As a result, the model can effectively capture abnormal motion patterns, such as reduced movement amplitude or impaired coordination of the upper limbs and torso, while maintaining lower computational complexity and improved generalization.

In the auditory branch, the mel-spectrogram representations are processed using a pretrained AST model to extract high-level auditory features that characterize speech impairments associated with stroke symptoms. The outputs of all modality encoders are subsequently mapped through projector layers to a shared embedding space with a unified dimensionality of 256. 
Formally, let $X_f$, $X_p$, and $X_s$ denote the input sequences for the face, pose, and speech modalities. The modality encoders extract feature representations as follows:

% \begin{equation}
% H_f = \text{AST}(X_f)
% \end{equation}

% \begin{equation}
% H_p = \text{MLPMixer}(X_p)
% \end{equation}

% \begin{equation}
% H_s = \text{AST}(X_s)
% \end{equation}
\begin{equation}
H_f = \text{AST}(X_f)
\end{equation}

\begin{equation}
H_p = \text{MLPMixer}(X_p)
\end{equation}

\begin{equation}
H_s = \text{AST}(X_s)
\end{equation}

The modality-specific features are then projected into a shared embedding space:

\begin{equation}
z_i = \text{Projector}(H_i), \quad i \in \{f,p,s\}
\end{equation}
where $z_f$, $z_p$, and $z_s$ denote the projected embeddings of the face, pose, and speech modalities, respectively. 
\subsection{Multimodal Fusion}

Once embeddings from the three modalities, namely facial landmarks, upper-body pose, and audio are obtained, the system performs multimodal fusion using a sigmoid-based gating mechanism,as illustrated in Fig.~\ref{Model}(c). Formally, let $z_f$, $z_p$, and $z_s$ denote the projected embeddings of the face, pose, and speech modalities, respectively. To adaptively modulate the contribution of each modality, a sigmoid-based gating mechanism is applied to each embedding.

\begin{equation}
z'_f = z_f \odot \sigma(z_f) + z_f
\end{equation}

\begin{equation}
z'_p = z_p \odot \sigma(z_p) + z_p
\end{equation}

\begin{equation}
z'_s = z_s \odot \sigma(z_s) + z_s
\end{equation}

where $\sigma(\cdot)$ denotes the sigmoid activation function and $\odot$ represents element-wise multiplication. This gating operation adaptively enhances informative modality features while suppressing less relevant signals.

The final multimodal representation is obtained by aggregating the gated embeddings:

\begin{equation}
z' = z'_f + z'_p + z'_s
\end{equation}

where $z'$ denotes the fused multimodal feature vector used for downstream classification.

The attention module plays a central role in coordinating and integrating information across modalities by adaptively learning the relative importance of each modality for the stroke screening task. Rather than employing simple fusion strategies such as feature concatenation or linear summation, the attention-based fusion enables the model to capture cross-modal relationships and emphasize discriminative and consistent cues while suppressing noisy or less informative signals.

In practice, the reliability and expressiveness of different modalities may vary across individuals and scenarios; for instance, facial asymmetry may be more pronounced in some cases, whereas speech impairment or motor dysfunction may provide stronger indicators in others. Through learned attention weights during training, the model dynamically adjusts its fusion strategy on a per-sample basis. Consequently, the final fused representation reflects a balanced and flexible interaction among facial expressions, body kinematics, and speech characteristics over time, leading to improved robustness and predictive performance in real-world prehospital settings.

\subsection{Classification Head}

% Finally, as shown in Fig.~\ref{Model}(d), the fused multimodal feature vector is fed into a classification head consisting of Layer Normalization, a GELU activation function, Dropout, and a Linear layer to perform binary classification between stroke and non-stroke conditions.

% The overall design of the proposed framework leverages the strengths of pretrained models while maintaining flexibility and scalability, making it suitable for practical deployment in prehospital and emergency care scenarios.

Finally, as shown in Fig.~\ref{Model}(d), the fused multimodal feature vector is fed into a classification head consisting of Layer Normalization, a GELU activation function, Dropout, and a Linear layer to perform binary classification between stroke and non-stroke conditions. 

Given the fused representation $z'$, the classification head first produces an intermediate feature representation through normalization and nonlinear transformation:

\begin{equation}
h = \text{GELU}(\text{LayerNorm}(z'))
\end{equation}

This representation is then passed through a linear classifier to compute the prediction score

\begin{equation}
y = \text{Linear(Dropout(}(h))
\end{equation}

and a sigmoid activation function is applied to obtain the final probability of the stroke class

\begin{equation}
\hat{y} = \text{Sigmoid}(y)
\end{equation}

The overall design of the proposed framework leverages the strengths of pretrained models while maintaining flexibility and scalability, making it suitable for practical deployment in prehospital and emergency care scenarios.
\section{Dataset} \label{Dataset}
\subsection{Data acquisition}
\begin{figure}
    \centering
    \includegraphics[width=0.9\linewidth]{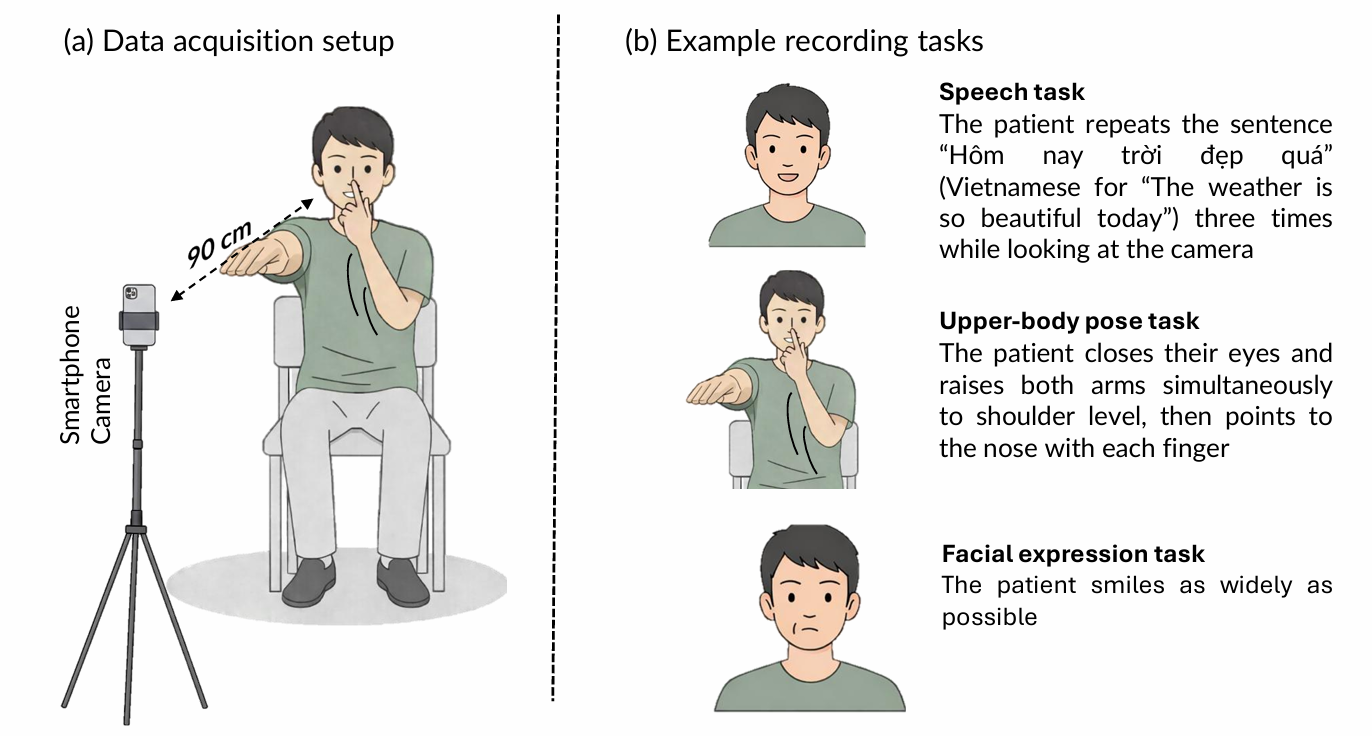}
    \caption{(a) Standardized data acquisition setup using a rear iPhone 12 camera mounted on a tripod at face height while the subject was seated. (b) Recording tasks including speech, upper-body pose, and facial expression assessment.}
    \label{setup}
\end{figure}

Data collection was conducted using the rear camera of an iPhone~12 mounted on a simple tripod, positioned at a height aligned with the subject’s face while seated. An overview of the recording setup and task-specific framing is illustrated in Fig. \ref{setup}. 
To ensure consistency and minimize environmental noise, all recordings were performed in a quiet indoor environment against a neutral background. Videos were captured in vertical orientation at a resolution of 1080p and a frame rate of 30~fps, with High Dynamic Range (HDR) disabled and mono audio recording enabled. Although not strictly required, the camera grid mode was activated to assist in centering the subject’s face within the frame. 
% Fig.~\ref{example} illustrates representative frames of the three video modalities captured in this study, including facial video, speech video, and upper-body movement video, corresponding to the tasks performed during the F.A.S.T. assessment.

% \begin{figure}
%     \centering
%     \includegraphics[width=1\linewidth]{Example_frame.pdf}
%     \caption{Illustration of the recording setup and example frames for the three video modalities used in the F.A.S.T. assessment.}
%     \label{example}
% \end{figure}

For the facial expression and speech tasks, the camera was positioned sufficiently close such that the face occupied approximately 80\% of the video frame, enabling detailed capture of facial movements. During the upper-body task, the camera was placed at a greater distance to ensure that both arms and hand movements remained fully visible throughout the recording, while minimizing excessive background space.

The dataset comprises recordings collected from 37 participants under a controlled prehospital setting. We define one multimodal sample (trial) as a triplet of recordings (Face, Speech, and Pose) captured for the same repetition. Each participant performed 6 repetitions (3 healthy + 3 simulated stroke), resulting in 6 multimodal samples per participant and 222 multimodal samples in total. Each multimodal sample contains three modality-specific videos, yielding 666 raw video clips $(222 \times 3)$  before quality filtering.  All participants performed a predefined set of tasks based on the F.A.S.T. assessment protocol. Detailed demographic information and data acquisition statistics are summarized in Table~\ref{dataset_summary}.

% \begin{table}[t]
% \centering
% \caption{Participant demographics and data acquisition details.}
% \label{dataset_summary}
% \renewcommand{\arraystretch}{1.25}
% \begin{tabular}{|p{3.8cm}|p{3.8cm}|}
% \hline
% \textbf{Category} & \textbf{Description} \\
% \hline
% Total participants & 37 \\
% \hline
% Female participants & 19 (51.35\%) \\
% \hline
% Male participants & 18 (48.65\%)\\
% \hline
% Age range (18--30) & 35 participants \\
% \hline
% Age range (30--40) & 2 participants \\
% \hline
% Assessment protocol & F.A.S.T. test \\
% \hline
% Number of task types & 3 \\
% \hline
% Task modalities & Face, Speech, Arm movement \\
% \hline
% Healthy trials per task & 3 repetitions \\
% \hline
% Simulated stroke trials per task & 3 repetitions \\
% \hline
% Total trials per task & 6 \\
% \hline
% Total recordings per participant & 18 videos \\
% \hline
% \end{tabular}
% \end{table}
\begin{table}[t]
\centering
\caption{Participant demographics and dataset statistics.}
\label{dataset_summary}
\renewcommand{\arraystretch}{1.25}
\begin{tabular}{|l|l|}
\hline
\textbf{Category} & \textbf{Description} \\
\hline
\multicolumn{2}{|l|}{\textbf{Participants}} \\
\hline
Total participants & 37 \\
Female / Male & 19 (51.35\%) / 18 (48.65\%) \\
Age (18--30) / (30--40) & 35 / 2 \\
\hline
\multicolumn{2}{|l|}{\textbf{Protocol and modalities}} \\
\hline
Assessment protocol & F.A.S.T.-inspired tasks \\
Task types & 3 (Face, Speech, Arm movement) \\
Modalities per sample & 3 (Face, Speech, Pose) \\
\hline
\multicolumn{2}{|l|}{\textbf{Recording counts}} \\
\hline
Repetitions per condition (per participant) & 3 healthy + 3 simulated stroke \\
Total repetitions (per participant) & 6 \\
Multimodal samples (per participant) & 6 \\
Total multimodal samples & 222 \\
Raw video clips (per participant) & 18 (6 samples $\times$ 3 modalities) \\
Total raw video clips & 666 (222 samples $\times$ 3 modalities) \\
\hline
\end{tabular}
\end{table}

As summarized in Table~\ref{fast_tasks}, each subject was instructed to perform three tasks corresponding to the components of the F.A.S.T. assessment. For the facial task, subjects were asked to smile as widely as possible under normal conditions. To simulate facial paralysis, they were instructed to smile using only one side of the face.

For the speech task, participants were required to repeat the sentence ``Hôm nay trời đẹp quá'', which translates to ``Today is such a nice day''. This sentence was selected due to its inclusion of multiple plosive phonemes, which are known to be particularly affected by dysarthria - a common speech impairment caused by disrupted neural control of oral muscles in stroke patients~\cite{asha_dysarthria}. When simulating stroke symptoms, subjects were asked to articulate the sentence more slowly and unclearly, mimicking impaired control of the mouth and tongue muscles.

For the upper-body task, participants were instructed to close their eyes, raise both arms to shoulder height, and subsequently touch the tip of their nose with each arm in turn. Although the original F.A.S.T. test does not include the nose touching component, this task was incorporated to assess motor coordination abilities, following the protocol proposed by Rodrigues et al.~\cite{rodrigues2017does}. To simulate stroke related motor deficits, subjects were instructed to mimic unilateral arm weakness, manifested as one arm hanging lower during elevation, accompanied by intentional hand tremors and imprecise nose touching movements.

\subsection{Data preprocessing}
In this study, facial landmarks were utilized to capture positional cues associated with stroke related symptoms, including facial drooping and asymmetrical mouth movements. By relying on landmark-based representations, the proposed approach emphasizes the spatial configuration of key facial regions while preserving participant privacy, as raw video frames are not processed directly. Facial landmarks were extracted on a frame-by-frame basis using Google’s MediaPipe library\cite{mediapipe_face_landmarker} with the Face Landmarker model, which outputs 478 landmarks defined by three dimensional coordinates. To specifically capture horizontal and vertical asymmetries, only the x and y coordinates were retained.

\begin{figure}[t]
    \centering
    \subfigure[Facial landmarks extracted using MediaPipe Face Landmarker\cite{mediapipe_face_landmarker} (478 landmarks).]{
        \includegraphics[width=0.45\linewidth]{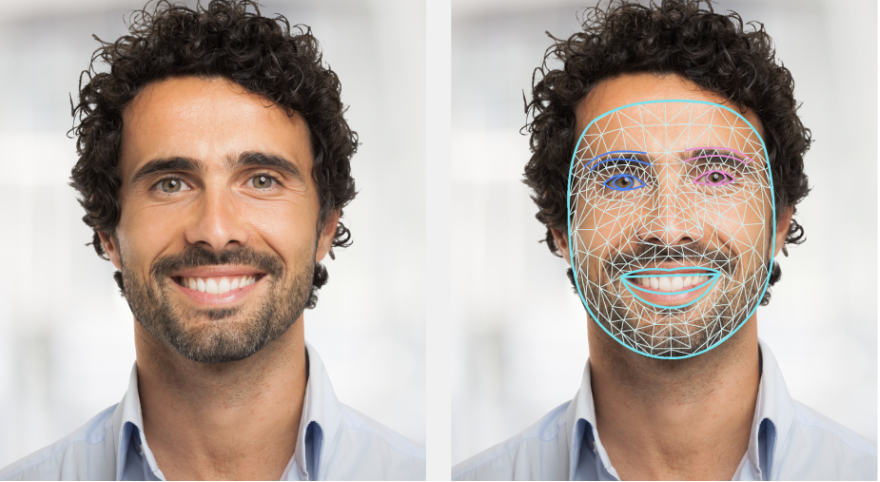}
        \label{face_landmarks}
    }
    \hfill
    \subfigure[Pose landmarks extracted using MediaPipe Pose Landmarker\cite{mediapipe_pose_landmarker} (33 landmarks).]{
        \includegraphics[width=0.45\linewidth]{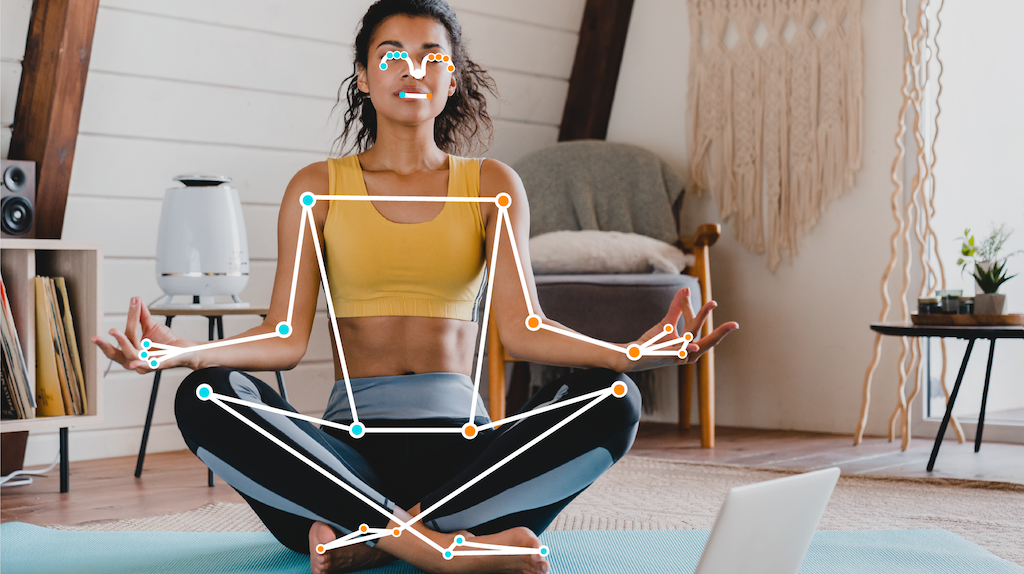}
        \label{pose_landmarks}
    }
    \caption{Illustrative examples of facial and pose landmarks adapted from the Google MediaPipe documentation.}
    \label{landmark_examples}
\end{figure}

Beyond facial features, body landmarks were extracted to characterize upper-body motion and skeletal dynamics, as body landmark representations have been shown to effectively capture motion patterns in video-based analysis of neurological conditions\cite{fleyeh2019extracting}. The MediaPipe Pose Landmarker\cite{mediapipe_pose_landmarker} model was employed, providing 33 landmarks with x, y, and z coordinates. In contrast to facial landmarks, all three axes were preserved for pose representation to retain depth information, which is particularly important for movements involving spatial displacement, such as finger to nose gestures. Collectively, these preprocessing pipelines transform each video into a temporal sequence of landmark coordinates, effectively representing motion as time-series data suitable for sequential modeling. Examples of the extracted facial and pose landmarks are illustrated in Figure~\ref{landmark_examples}.

\begin{figure}[t]
    \centering
    \subfigure[healthy]{
        \includegraphics[width=0.45\linewidth]{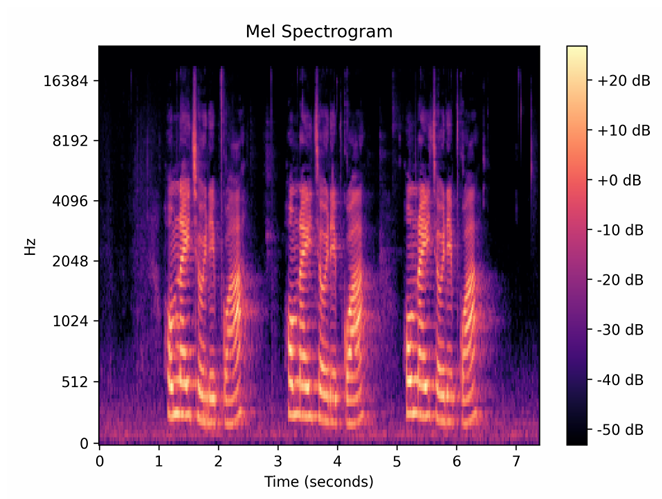}
        \label{healthy}
    }
    \hfill
    \subfigure[stroke-simulated]{
        \includegraphics[width=0.45\linewidth]{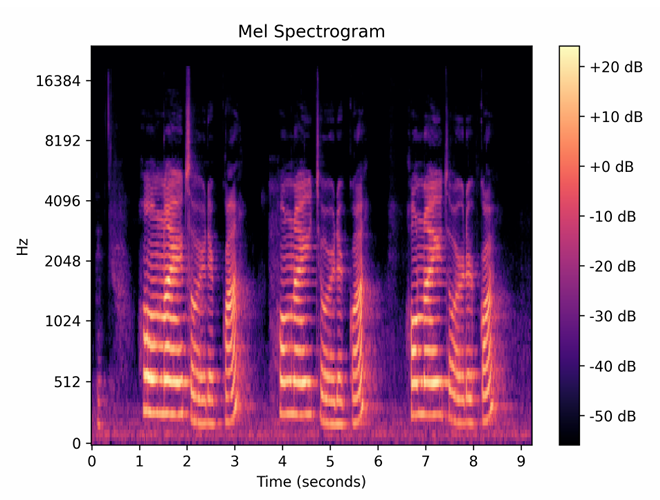}
        \label{pose_landmarks}
    }
    \caption{Side-by-side comparison of mel spectrograms from (a) a healthy subject and (b) a stroke-simulated subject, illustrating differences in temporal and spectral speech patterns at 80 mel bins.}
    \label{mel_compare}
\end{figure}

Audio signals were represented using mel spectrograms, which effectively capture time-frequency characteristics relevant to speech analysis\cite{zhu2023multiscale}. Each audio recording was first downsampled to 16 kHz and transformed using the Short-Time Fourier Transform, followed by a mel-scale mapping that emphasizes lower speech frequencies. This representation highlights articulatory differences between vowels and consonants, which are particularly important for identifying dysarthria-related speech impairments. To ensure uniform input dimensions, all mel spectrograms were generated with 80 mel bins and temporally padded or truncated to a fixed length of 256 frames. Fig.~\ref{mel_compare} provides a qualitative comparison between healthy and stroke simulated speech, highlighting observable differences in temporal frequency structures captured by mel spectrograms.

\begin{table}
\centering
\caption{Tasks used in the F.A.S.T. assessment and their corresponding stroke-related impairments.}
\label{fast_tasks}
\resizebox{\columnwidth}{!}{%
\begin{tabularx}{\columnwidth}{|X|X|X|}
\hline
\textbf{Task} &
\textbf{Expected behavior of stroke-affected patients} &
\textbf{Physical / vocal impairment} \\
\hline
The patient smiles as widely as possible &
Facial movement is uneven between the two sides of the face &
Facial drooping \\
\hline
The patient repeats the sentence ``Hôm nay trời đẹp quá'' (Vietnamese for ``The weather is so beautiful today'') three times while looking at the camera &
Speech becomes slowed and pronunciation is unclear &
Slurred speech \\
\hline
The patient closes their eyes and raises both arms simultaneously to shoulder level, then points to the nose with each finger &
One arm does not lift as quickly or as high as the other. When pointing to the nose, the finger drifts slightly and may not reach the nose on the first attempt &
Arm weakness and coordination problems \\
\hline
\end{tabularx}}
\end{table}

\section{Experiments} \label{Experiment}
\subsection{Implementation Details}

The proposed system consists of multiple modality-specific branches designed to process heterogeneous input signals. Facial landmark sequences were modeled using a Transformer encoder initialized from a pretrained Audio Spectrogram Transformer (AST) backbone with a base configuration. Speech signals were processed using an AST model with a tiny configuration to extract discriminative acoustic representations from mel spectrograms. For upper-body pose data, an MLP-Mixer architecture was employed and trained from scratch to capture temporal and spatial interactions among pose landmarks.

All models were implemented in PyTorch and trained for 200 epochs using the Adam optimizer with the Binary Cross-Entropy loss function. A decision threshold of 0.5 was applied to classify samples as stroke or non-stroke. The experiments were conducted on an NVIDIA GeForce RTX 3090 GPU. 

After data cleaning, the dataset consisted of 222 valid video samples and was split in a subject-independent manner into training, validation, and test sets. Specifically, 29 subjects (116 videos) were used for training, while 4 subjects (24 videos) were allocated to each of the validation and test sets, ensuring no subject overlap across splits.
Model performance was evaluated using Accuracy, Area Under the Receiver Operating Characteristic Curve (AUC), and the balanced F1-score. Table~\ref{tab:config} summarizes the main model configurations and training hyperparameters used in our experiments.
\begin{table}[t]
\centering
\caption{Model configuration and training hyperparameters}
\label{tab:config}
\begin{tabular}{lc}
\toprule
\midrule
\textbf{Parameter} & \textbf{Value} \\
\midrule
Face encoder & AST (base) \\
Speech encoder & AST (tiny) \\
Pose encoder & MLP-Mixer \\
Embedding dimension & 256 \\
Optimizer & Adam \\
Learning rate & $1\times10^{-4}$ \\
Batch size & 8 \\
Training epochs & 200 \\
Loss function & Binary Cross-Entropy \\
Decision threshold & 0.5 \\
Framework & PyTorch \\
GPU & NVIDIA RTX 3090 \\
\midrule
\bottomrule
\end{tabular}
\end{table}
% \subsection{Experimental Result}
\subsection{Computational Efficiency and Inference Cost}
To address practical deployment considerations, we report the model size and inference efficiency of the proposed multimodal framework. Table~\ref{tab:efficiency} summarizes the parameter breakdown across modality-specific encoders and the fusion head. The model consists of three modality-specific branches and a lightweight fusion head. 
The face branch encodes temporal facial landmark sequences using a Transformer-based encoder (initialized from a pretrained AST encoder), which provides strong representation capacity but also contributes the majority of parameters. 
The speech branch converts raw audio into mel-spectrograms and applies an AST-tiny encoder to capture time-frequency patterns related to dysarthria; this branch is substantially smaller than the face encoder. 
The pose branch processes upper-body pose landmark sequences with an MLP-Mixer, offering a compact and efficient alternative to attention-heavy architectures while still modeling spatiotemporal motion cues. 
Finally, the fusion head aggregates the three embeddings via attention and produces the binary prediction with negligible parameter overhead. The overall model contains 92.490M parameters (352.82 MB when stored in 32-bit floating point), where the facial branch contributes the largest portion of parameters. In addition, we measure the end-to-end inference latency of the full multimodal pipeline at 7.19 ms per sample (batch size $=4$ during measurement). These results indicate that the proposed architecture can support real-time screening requirements, while also highlighting that future optimization efforts, such as model compression or quantization, can primarily focus on the facial encoder to further reduce the memory footprint.

\begin{table}[t]
\centering
\caption{Efficiency summary of the proposed model.}
\label{tab:efficiency}
\renewcommand{\arraystretch}{1.15}
\begin{tabular}{lr}
\toprule
\midrule
\textbf{Item} & \textbf{Value} \\
\midrule
Face encoder params  & 85.791 M \\
Audio encoder params & 5.809 M \\
Pose encoder params  & 0.643 M \\
Fusion + head params & 0.247 M \\
\midrule
\textbf{Total params} & \textbf{92.490 M} \\
Model size (params only) & 352.82 MB \\
Inference time & 7.19 ms / sample \\
\midrule
\bottomrule
\end{tabular}
\end{table}

\subsection{Performance Comparison Across Model Variants}

The results in the table \ref{results} reveal a clear performance gap between unimodal and multimodal models, while also highlighting the different levels of contribution of each modality in the stroke screening task. Among the unimodal approaches, the pose-only model achieves the lowest performance across all metrics, with an Accuracy of 0.6667 and an F1-score of 0.6364. This indicates that upper-body pose information, when used in isolation, provides only limited motion-related cues and is insufficient to reliably distinguish between stroke and non-stroke conditions in the prehospital setting. The voice-only model shows a substantial improvement over the pose-only baseline, achieving both Accuracy and F1-score values of 0.8333. This suggests that speech signals contain more discriminative features associated with neurological impairments, particularly symptoms such as slurred or slowed speech. However, the Specificity of this model is limited to 0.8333, indicating the presence of a non-negligible number of false-positive predictions when relying solely on audio information. In contrast, the face-only model yields the highest performance among the unimodal approaches, with an Accuracy of 0.8750 and an AUC of 0.9306. Notably, this model achieves a Specificity of 1.0000, demonstrating its strong ability to correctly identify non-stroke cases. Nevertheless, its Sensitivity remains at 0.7500, implying that some stroke cases are still missed when relying exclusively on facial expressions and asymmetry cues.

In addition to unimodal models, we evaluate pairwise fusion configurations to analyze the individual contributions and complementarity of different modalities. The results show that all pairwise fusion models outperform their unimodal counterparts; however, their effectiveness varies substantially depending on the modality combination. The Face + Pose model achieves an Accuracy of 0.8750 and an F1-score of 0.8571, with an AUC of 0.9375 and a Specificity of 1.0000, while its Sensitivity remains limited at 0.7500. This indicates that upper-body pose provides only limited complementary information for fully detecting stroke cases. In contrast, the Voice + Pose model exhibits a more balanced performance, with an F1-score of 0.8696 and an AUC of 0.9735, highlighting the added value of speech information in improving sensitivity. Notably, the Face + Voice configuration achieves the best performance among the pairwise fusion models, with an Accuracy of 0.9167, an AUC of 1.0000, and an F1-score of 0.9091. These results suggest that facial appearance and speech are the two most discriminative modalities in the F.A.S.T. assessment, and their combination effectively captures the key stroke-related symptoms even without incorporating all three modalities.

When multimodal information is integrated, the fusion model with frozen pretrained weights exhibits a clear improvement in overall discriminative capability, with the AUC increasing to 0.9792 and the F1-score reaching 0.8571. Freezing the pretrained weights allows the model to preserve stable feature representations while mitigating overfitting, which is particularly important given the limited size of the training data. However, the Sensitivity of this model remains at 0.7500, suggesting that stroke case detection is not yet fully optimized. The best overall performance is achieved by the fusion model with fine-tuned pretrained weights, which attains an Accuracy of 0.9583, a perfect AUC of 1.0000, and an F1-score of 0.9600. Importantly, this model reaches a Sensitivity of 1.0000, indicating that all stroke cases in the test set are correctly identified, while maintaining a high Specificity of 0.9167. These results demonstrate that fine-tuning the pretrained weights during multimodal fusion enables the model to better capture the complementary interactions among facial, speech, and motor cues, leading to a substantial improvement in overall performance.

To further evaluate the effectiveness of the proposed multimodal framework, we additionally compare our method with two commonly used multimodal baselines: Late Fusion and Feature Concatenation followed by a Multilayer Perceptron (Concat + MLP). The Late Fusion approach aggregates the predictions of the unimodal models by averaging their output probabilities. As shown in Table \ref{results}, this baseline achieves an Accuracy of 0.9167 and an F1-score of 0.9091, which is comparable to the best pairwise fusion configuration (Face + Voice). This result indicates that combining predictions from independently trained unimodal models can already capture complementary information across modalities to a certain extent. The Concat + MLP baseline, which directly concatenates the feature representations from the three modalities and feeds them into a multilayer perceptron for classification, achieves an Accuracy of 0.8750 and an F1-score of 0.8571. Although this approach allows the classifier to access multimodal information simultaneously, the simple concatenation strategy lacks an explicit mechanism to model complex cross-modal interactions. As a result, its performance remains comparable to some pairwise fusion configurations but does not fully exploit the complementary relationships among facial, speech, and pose features.

Overall, the experimental results confirm the clear advantage of multimodal learning over unimodal approaches, particularly in achieving a balanced trade-off between Sensitivity and Specificity. Furthermore, the effective use of pretrained models combined with an appropriate fine-tuning strategy plays a crucial role in enhancing the accuracy and reliability of stroke screening systems in prehospital environments, where data are typically limited and subject to noise.
\begin{table}
\centering
\caption{Performance comparison of different models. The best results are highlighted in bold.}
\label{results}
\renewcommand{\arraystretch}{1.3}

\resizebox{\columnwidth}{!}{%
\begin{tabular}{lccccc}
\toprule
\midrule
\textbf{Model} & \textbf{Accuracy} & \textbf{AUC} & \textbf{F1-score} & \textbf{Sensitivity} & \textbf{Specificity} \\
\midrule
Pose only & 0.6667 & 0.8125 & 0.6364 & 0.5833 & 0.7500 \\
% \hline
Voice only & 0.8333 & 0.8819 & 0.8333 & 0.8333 & 0.8333 \\
% \hline
Face only & 0.8750 & 0.9306 & 0.8750 & 0.7500 & 1.0000\\

% \hline
Face + Pose & 0.8750 & 0.9375 & 0.8571 & 0.7500 & 1.0000\\
% \hline
Voice + Pose & 0.8750 & 0.9735 & 0.8696 & 0.8333 & 0.9167 \\
% \hline
Face + Voice & 0.9167 & 1.0000 & 0.9091 & 0.8333 & 1.0000 \\
% \hline
 Late fusion & 0.9167 &  0.9861 &  0.9091 &  0.8333 & 1.0000\\

Concat + MLP &  0.8750 & 0.9583 & 0.8571 &  0.8333 &  0.9167\\
Fusion (Frozen weight) & 0.8750 & 0.9792 & 0.8571 & 0.7500 & \textbf{1.0000 }\\
% \hline
\textbf{Fusion (Fine-tuned) (Ours)} & \textbf{0.9583} &\textbf{ 1.0000} & \textbf{0.9600} & \textbf{1.0000 }& 0.9167 \\
\midrule
\bottomrule
\end{tabular}}
\end{table}
\section{Ablation Study} \label{Ablation Study}
\subsection{Performance between difference fusion strategies}
Table \ref{ablation} presents a performance comparison of different fusion strategies to evaluate the impact of each feature aggregation method within the multimodal framework. In particular, we compare the proposed approach with several commonly used multimodal fusion baselines, including feature concatenation, simple summation, and learnable weighted summation. Overall, learnable fusion strategies consistently outperform simple concatenation, indicating that explicitly modeling the relationships among modalities plays a crucial role in prehospital stroke screening.

The Concat method achieves an Accuracy of 0.8750 and an F1-score of 0.8696, with a relatively high AUC of 0.9792. While this indicates strong overall discriminative capability, the method remains limited in simultaneously optimizing Sensitivity and Specificity. The Sum strategy yields comparable Accuracy (0.8750) and achieves a perfect AUC of 1.0000; however, its F1-score and Sensitivity decrease to 0.8571 and 0.7500, respectively, suggesting that simple linear aggregation lacks sufficient flexibility to capture positive stroke cues effectively.

In contrast, the Learnable Weighted Sum strategy shows a clear improvement, achieving an Accuracy of 0.9167 and an F1-score of 0.9091 while maintaining an AUC of 1.0000. These results demonstrate that learning modality-specific contribution weights enables better adaptation to the data and enhances discriminative performance. The highest performance is achieved by Attention Fusion, which attains an Accuracy of 0.9583 and an F1-score of 0.9600, together with a Sensitivity of 1.0000, indicating complete detection of stroke cases in the test set.

Overall, the ablation results confirm that learnable fusion strategies particularly attention-based fusion provide substantial advantages over simpler aggregation methods. This finding highlights the importance of modeling cross-modal interactions to improve both performance and reliability in prehospital stroke screening systems.
\begin{table}
\centering
\caption{Performance Comparison of Different Fusion Strategies}
\label{ablation}
\renewcommand{\arraystretch}{1.3}

\resizebox{\columnwidth}{!}{%
\begin{tabular}{lccccc}
\toprule
\midrule
\textbf{Fusion Strategy} & \textbf{Accuracy} & \textbf{AUC} & \textbf{F1-score} & \textbf{Sensitivity} & \textbf{Specificity} \\

\midrule
Concat & 0.8750 & 0.9792 & 0.8696 & 0.8333 & 0.9167 \\
% \hline
Sum & 0.8750 & 1.0000 & 0.8571 & 0.7500 & \textbf{1.0000} \\
% \hline
Learnable Weighted Sum & 0.9167 & 1.0000  & 0.9091 & 0.8333 & \textbf{1.0000} \\
% \hline
\textbf{Attention Fusion (Ours)} & \textbf{0.9583} & \textbf{1.0000} & \textbf{0.9600} & \textbf{1.0000} & 0.9167 \\
\midrule
\bottomrule
\end{tabular}}
\end{table}
\subsection{Performance under controlled modality corruptions}
\begin{table}
\centering
\caption{Performance under different augmentation settings. A checkmark indicates the modality is augmented.}
\label{tab:augment_env}
\renewcommand{\arraystretch}{1.2}

\begin{tabular}{ccc|ccccc}
\toprule
\midrule
\multicolumn{3}{c|}{\textbf{Corrupted Modality}} &
\multicolumn{5}{c}{\textbf{Performance}} \\
\cmidrule(lr){1-3} \cmidrule(lr){4-8}
% \cmidrule
\textbf{Face} & \textbf{Pose} & \textbf{Speech }& \textbf{Acc} & \textbf{AUROC} & \textbf{F1} & \textbf{Sen} & \textbf{Spec} \\
\midrule
&  &  &0.9583 & 1.0000 & 0.9600 & 1.0000 & 0.9167 \\
$\checkmark$ &  &  & 0.9167 & 0.9861 & 0.9231 & 1.0000 & 0.8333 \\
 & $\checkmark$ &  & 0.9583 & 1.0000 & 0.9600 & 1.0000 & 0.9167 \\
 &  & $\checkmark$ & 0.8750 & 0.9722 & 0.8889 & 1.0000 & 0.7500 \\
$\checkmark$ & $\checkmark$ &  & 0.9167 & 0.9861 & 0.9231 & 1.0000 & 0.8333 \\
 & $\checkmark$ & $\checkmark$ & 0.8750 & 0.9722 & 0.8889 & 1.0000 & 0.7500 \\
$\checkmark$ &  & $\checkmark$ & 0.8750 & 0.9583 & 0.8889 & 1.0000 & 0.7500 \\
$\checkmark$ & $\checkmark$ & $\checkmark$ & 0.8333 & 0.9514 & 0.8571 & 1.0000 & 0.6667 \\
\midrule
\bottomrule
\end{tabular}
\end{table}
To assess robustness under challenging prehospital recording conditions, we perform a controlled \emph{test-time} corruption study where each modality is corrupted while the remaining modalities are kept clean, and we additionally consider multi-modality corruptions. For Face and Pose, corruptions are applied at the raw video-frame level and landmarks are then re-extracted using the same MediaPipe pipeline. Specifically, we apply one of three visual perturbations to simulate common field conditions: (i) motion (random affine transformations to mimic handheld camera motion), (ii) outdoor (photometric changes via color jitter to emulate illumination/exposure variations), and (iii) blur (Gaussian blur to simulate defocus or motion blur). For Speech, we inject additive noise directly into the raw waveform at a fixed SNR (10dB) prior to mel-spectrogram computation, approximating moderate background noise. As reported in Table~\ref{tab:augment_env}, the proposed multimodal model preserves perfect sensitivity across all corruption settings (Sen = 1.0000), which is critical for screening. Under single-modality corruptions, Pose corruption yields the smallest drop (Acc = 0.9583, F1 = 0.9600), Face corruption causes a moderate decrease (Acc = 0.9167, F1 = 0.9231), while Speech corruption shows the largest degradation among single-modality cases (Acc = 0.8750, F1 = 0.8889). When multiple modalities are corrupted simultaneously, performance decreases as expected; the most adverse setting (Face+Pose+Speech) results in Acc = 0.8333 and F1 = 0.8571, with the reduction mainly reflected in specificity (Spec = 0.6667), i.e., more false positives rather than missed positives, which is a preferable trade-off for a screening-oriented application.
\subsection{Pose Branch Temporal-Segment Ablation}
\label{subsec:pose_temporal_ablation}

To further analyze the pose modality, we conduct a temporal-segment ablation on the upper-body
pose landmark sequences. For each sample, we extract a fixed-length window of $T=200$ frames
using three strategies: early (first $T$ frames), center (a centered window), and
late (last $T$ frames). All other settings are kept unchanged.

Table~\ref{tab:pose_temporal_ablation} shows that the temporal window selection affects the
pose-only performance. The center window yields the most balanced results, achieving
Acc $=66.67\%$ and F1 $=0.6364$ with Sen $=0.5833$ and Spec $=0.7500$. The early window
produces higher specificity (Spec $=0.9167$) but substantially lower sensitivity (Sen $=0.2500$),
indicating more missed positive cases when only the initial segment is used. The late window
achieves perfect specificity (Spec $=1.0000$) but lower sensitivity (Sen $=0.3333$), suggesting that
while late-stage movements can be discriminative, relying only on the final segment is insufficient
for consistent positive detection. Overall, these results indicate that coordination cues in the
nose-touching task are not uniformly distributed over time, and the pose modality is sensitive to
temporal cropping.
\begin{table}[t]
\centering
\caption{Pose-only performance under different temporal window selections (window length $T=200$ frames).}
\label{tab:pose_temporal_ablation}
\renewcommand{\arraystretch}{1.15}
\begin{tabular}{lccccc}
\toprule
\textbf{Window} & \textbf{Acc} & \textbf{AUROC} & \textbf{F1} & \textbf{Sen} & \textbf{Spec} \\
\midrule
Early  & 0.5833 & 0.7778 & 0.3750 & 0.2500 & 0.9167 \\
Center & \textbf{0.6667} & \textbf{0.8125} & \textbf{0.6364} & \textbf{0.5833} & \textbf{0.7500} \\
Late   & 0.6667 & 0.8333 & 0.5000 & 0.3333 & 1.0000 \\
\bottomrule
\end{tabular}
\vspace{0.4em}

\begin{tabular}{lcccc}
\toprule
\textbf{Window} & \textbf{TP} & \textbf{FP} & \textbf{FN} & \textbf{TN} \\
\midrule
Early  & 3  & 1 & 9 & 11 \\
Center & 7  & 3 & 5 & 9  \\
Late   & 4  & 0 & 8 & 12 \\
\bottomrule
\end{tabular}
\end{table}

\section{Dicussion} \label{Dicussion}
Overall, these results highlight the potential of multimodal learning combined with transfer learning in data-limited prehospital settings. Achieving a Sensitivity of 1.0000 with the fine-tuned fusion model is particularly important for early stroke screening, where minimizing missed cases is critical. At the same time, the high Specificity demonstrates reliable control of false alarms, suggesting that the proposed approach is promising as a screening-oriented decision-support prototype.

\subsection{Simulated vs.\ Clinical Data}
Despite these promising results, several limitations related to data scale and clinical representativeness remain. The dataset is relatively small, consisting of 222 samples from a limited number of participants, which may restrict generalization to more diverse prehospital scenarios and increase the risk of overfitting, particularly for multimodal models with many parameters. To mitigate this limitation, we leverage transfer learning by initializing the visual and audio encoders from pretrained models, which helps improve representation learning under limited data conditions. In addition, data augmentation strategies were applied during training and analyzed in the ablation study to improve model robustness. In addition, the data were collected using simulated stroke symptoms rather than recordings from real patients; therefore, the current study should be regarded as a proof-of-concept rather than a clinically validated, deployable screening system. While simulated data enable controlled acquisition and balanced labeling, they may not fully reflect the variability and complexity of real-world stroke presentations (e.g., differences in symptom severity, mixed symptom profiles, comorbidities, and older demographics). Moreover, real prehospital recordings can involve uncontrolled conditions such as camera motion, occlusions, and background noise, which may introduce distribution shifts. Accordingly, validation on larger, independent datasets, ideally including external test sets and prospective evaluations on real patients, is required before deployment.

\subsection{Demographic Bias and Age Generalizability}
Our cohort is skewed toward younger participants (35/37 are aged 18--30), whereas stroke incidence is substantially higher in older populations. This age imbalance limits the current applicability of our findings, as age can change the baseline characteristics of cues used in F.A.S.T.-style screening. For example, facial geometry and habitual asymmetry may vary with age (e.g., sagging and reduced muscle tone), speech can be affected by age-related changes and comorbidities, and upper-body motion patterns may be influenced by reduced mobility or musculoskeletal conditions, potentially increasing false positives or masking subtle stroke-like deficits. Consequently, generalization to the target older population remains unproven in this study. In future work, we will expand data collection to include older participants and real patients, and perform age-stratified evaluation (and, if needed, calibration or domain adaptation) to ensure robust performance across age groups.

\section{Conclusion} \label{Conclusion}
In this study, we propose a multimodal learning based approach for prehospital stroke screening that jointly exploits information from facial appearance, speech, and upper-body movements acquired during the F.A.S.T. assessment. The system is designed to be non-invasive, easy to deploy, and well suited to prehospital scenarios, where early detection and rapid decision support are critical.

Experimental results demonstrate that the proposed multimodal model consistently outperforms unimodal baselines across most evaluation metrics, particularly in achieving a favorable balance between Sensitivity and Specificity. Leveraging pretrained models in combination with an attention-based fusion strategy substantially improves overall performance while maintaining training stability under limited data conditions. Notably, the proposed method achieves complete detection of stroke cases in the test set, highlighting its potential for practical deployment in early stroke screening.

Despite limitations related to dataset size and representativeness, the obtained results confirm the feasibility of applying multimodal deep learning to prehospital stroke screening. Future work will focus on expanding the dataset with recordings from real stroke patients, diversifying data acquisition scenarios, and evaluating the proposed system on independent datasets to further enhance its generalization capability and reliability.

\section*{Acknowledgment}
 The authors would like to sincerely thank the Vietnam Young Talent Support Fund, Tan Hiep Phat Trading – Service Co., Ltd., and the Ben Dam Me Award Fund for their valuable support and encouragement of this work.

\bibliographystyle{IEEEtran} % sorted IEEE style
\bibliography{template.bib} % name your BibTeX database
% \hfill {\it Received on October 14, 2003}

% \hfill {\it Accepted on January 14, 2004}
\end{document}